\documentclass[9pt,conference,final]{IEEEtran}
\IEEEoverridecommandlockouts
\usepackage[symbol]{footmisc}
\usepackage{cite}
\usepackage{amsmath,amssymb,amsfonts}
\usepackage{algorithmic}
\usepackage{graphicx}
\usepackage{textcomp}
\usepackage{xcolor}
\usepackage[switch]{lineno}
\usepackage{fancyhdr}
\usepackage{bm}
\usepackage{multirow}
\usepackage{booktabs}
\usepackage{float}
\usepackage[inkscapelatex=false]{svg} 
\usepackage{epstopdf}

\def\BibTeX{{\rm B\kern-.05em{\sc i\kern-.025em b}\kern-.08em
    T\kern-.1667em\lower.7ex\hbox{E}\kern-.125emX}}

\begin{document}

\pagestyle{fancy}
\pagenumbering{gobble}


\title{SDG-L: A Semiparametric Deep Gaussian Process based Framework for Battery Capacity Prediction}

\author{%
  \IEEEauthorblockN{%
    Hanbing Liu\IEEEauthorrefmark{1},
    Yanru Wu\IEEEauthorrefmark{1},
    Yang Li\IEEEauthorrefmark{2},
    Ercan E. Kuruoglu\IEEEauthorrefmark{2} and
    Xuan Zhang
  }%
  \IEEEauthorblockA{\textit{Tsinghua-Berkeley Shenzhen Institute, Tsinghua Shenzhen International Graduation School, Tsinghua University}}
  \IEEEauthorblockA{\IEEEauthorrefmark{2} Corresponding authors. E-mail: yangli@sz.tsinghua.edu.cn, kuruoglu@sz.tsinghua.edu.cn}
  \IEEEauthorblockA{\IEEEauthorrefmark{1} These authors contributed equally to this work.}
}

\maketitle

\begin{abstract}
Lithium-ion batteries are becoming increasingly omnipresent in energy supply. However, the durability of energy storage using lithium-ion batteries is threatened by their dropping capacity with the growing number of charging/discharging cycles. An accurate capacity prediction is the key to ensure system efficiency and reliability, where the exploitation of battery state information in each cycle has been largely undervalued. In this paper, we propose a semiparametric deep Gaussian process regression framework named SDG-L to give predictions based on the modeling of time series battery state data. By introducing an LSTM feature extractor, the SDG-L is specially designed to better utilize the auxiliary profiling information during charging/discharging process. In experimental studies based on NASA dataset, our proposed method obtains an average test MSE error of 1.2\textperthousand. We also show that SDG-L achieves better performance compared to existing works and validate the framework using ablation studies.

\end{abstract}


\begin{figure*}[h]
	\begin{center}
		\centerline{\includegraphics[width=2\columnwidth]{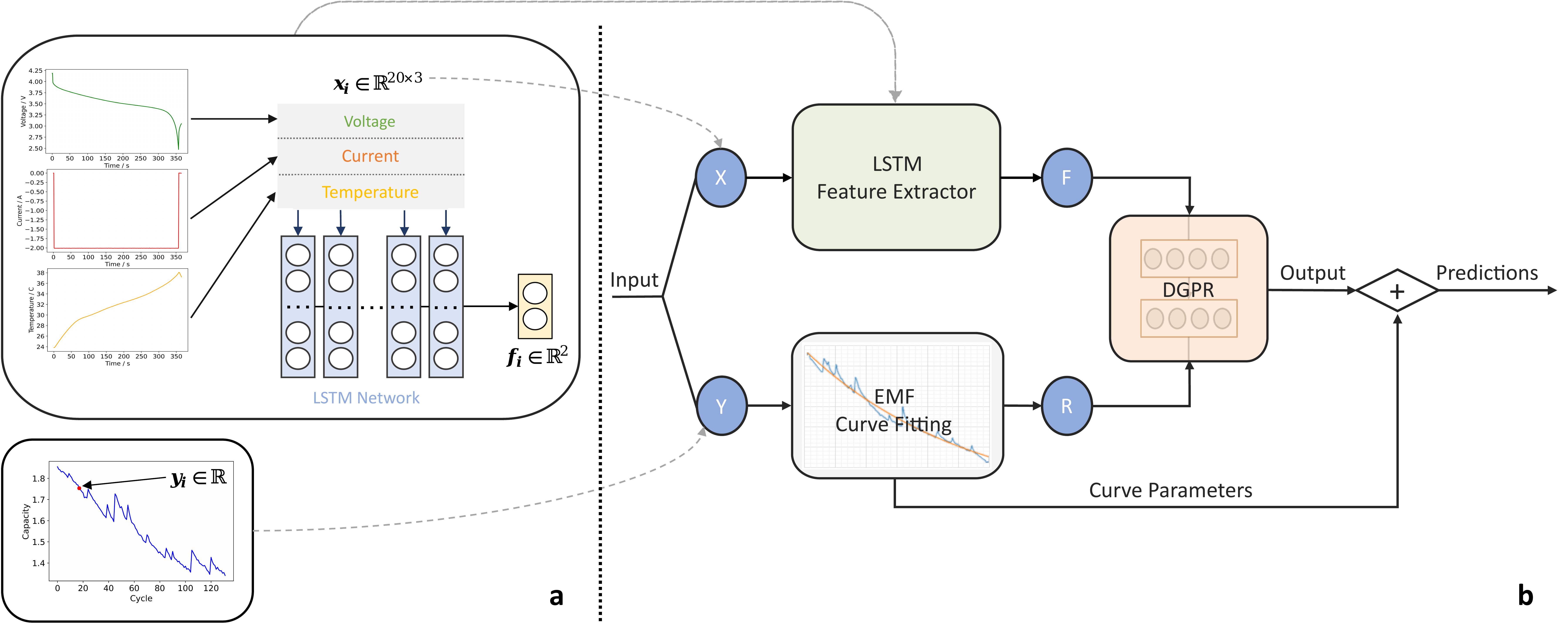}}
		\caption{\textbf{Method Framework.} Part b is the pipeline of our method and the detail of our framework components are in part a. In our framework, X data (input exogenous information) and Y data (battery capacity) are respectively processed to features (F) and residuals (R) before being fed into the DGPR module. X is put through an LSTM feature extractor (illustrated in part a) to get features (F), while residuals (R) are derived from the EMF curve fitting on Y. Predictions can be calculated by adding the DGPR output and EMF results for the test set using the fitted curve parameters. Specifically, the time series of voltage, current, and temperature observed at each cycle is embedded into a $20\times3$ matrix and fed to an LSTM extractor, whose output feature dimension is set to be 2. The LSTM extractor and the DGPR module will be trained simultaneously.}
		\label{framework}
	\end{center}
 \vspace{-0.5cm}

\end{figure*}

\section{Introduction}

From electric vehicles to stationary storage applications, lithium-ion batteries (LIBs) are widely used as energy storage systems. However, the growing cell number and energy density of lithium-ion batteries come with a greater risk of catastrophic events, where an insufficient knowledge of their health state or useful life can expose the system to a sudden drop in performance or keep it running in a dangerous manner\cite{cabrera2016calculation}. As a result, LIBs are frequently oversized and under-utilized to ensure system reliability\cite{richardson2017gaussian}. The alleviation of such inefficiency particularly relies on the prediction of battery health state\cite{neubauer2011ability}.

Typically, the main concern of battery health lies in capacity, which notably drops with the growth of charging/discharging cycles and is challenging to precisely predict. Many methods have been adapted to the capacity prediction problem due to its significance and have achieved remarkable performances\cite{zhang2017overview}. Conventional approaches typically rely on degradation modeling via electrochemical or equivalent circuit models\cite{wang2011cycle}, requiring plentiful knowledge of battery mechanism. Meantime, given the time series nature of the capacity prediction problem and the rapid development of data analytics, data-driven approaches are becoming the mainstream of battery health estimation\cite{li2019data}. These methods include parametric methods such as curve fitting\cite{rong2006analytical}, support vector machines\cite{patil2015novel} and different types of neural networks (\cite{you2016real}-\cite{how2020state}), as well as non-parametric methods such as k-nearest neighbours\cite{hu2014data}, XGBoost\cite{ma2021XGBoost}, Bayesian techniques\cite{saha2008uncertainty}, Monte Carlo method\cite{he2011prognostics} and Gaussian process regression (GPR)\cite{richardson2018gaussian}.

Despite the remarkable modeling capability of large parametric models as derived in \cite{wang2013prognostics} and \cite{hong2020towards}, these methods are limited in assuming that a specific deterministic functional form should exist for mapping available information to the desired prediction. On the contrary, stochastic models permit an expressivity that not only predicts future values of the variable of interest but also expresses the uncertainty associated with these values\cite{richardson2017gaussian}, providing additional information for decision making\cite{klas2018uncertainty}. Among them, GPR as a Bayesian method with solid theory support\cite{williams2006gaussian} has gained special attention from researchers these years and displayed a certain superiority in experimental studies\cite{liu_hu_wei_li_jiang_2019}. Studies in \cite{liu2013prognostics} and \cite{richardson2017gaussian} both successfully applied different kernelized GPR to battery capacity prediction, and showed that their predictive accuracy was improved when a linear or quadratic explicit mean function (EMF) was used, where the algorithm actually becomes \textit{semiparametric}, referring to the combination of parametric EMF model and non-parametric GPR method. Being non-parametric and probabilistic, GPR displays extraordinary performance in many problems, yet its expressivity is still constrained compared to deep models with artificial neural networks (ANNs). This limitation is solved by the construction of deep Gaussian process regression (DGPR) by stacking Gaussian process based layers together into a network structure\cite{damianou2013deep}. 

Moreover, an important yet highly undervalued aspect of battery data analysis is the mining of state data during every charging/discharging cycle. Typically, a battery management system (BMS) monitors voltage, current and temperature change in each charging/discharging process\cite{li2019data}, deriving an input of three-channel discharging profile data for each cycle. Such high dimensional data is intractable for most algorithms without the help of effective feature extractors. Tagade et al. \cite{tagade2020deep} succeeded in using ANN for feature extraction, while we assume that an LSTM suits more the time series information. 

Therefore, the methodology we propose is a \underline{S}emiparametric model based on the combination of \underline{D}eep \underline{G}aussian process and \underline{L}STM (SDG-L), with the latter acting as the feature extractor for profiling information. The whole framework and its experimental performance will be revisited later in the corresponding sections. The main contribution of this work lies in three aspects:
\begin{itemize}
\item[1)] We formulate the problem of utilizing the profiling information in each charging/discharging cycle for capacity prediction.
\item[2)] We introduce an LSTM feature extractor into our model framework to exploit the time series properties of BMS state data, cascading it with DGPR and an EMF curve fitting to leverage both the prior knowledge of battery capacity degradation nature and the higher representative ability of DGPR compared to plain GPR and parametric methods.
\item[3)] We validate our framework with experiments conducted on the NASA dataset using ablation study, which also derives good performance compared to existing methods. 
\end{itemize}

The rest of this paper is organized as follows. In section \ref{sec:background}, we will provide a brief introduction to the basic techniques used in our framework. Then, a thorough description of the problem definition and our proposed method will be presented in section \ref{sec:methodology}; experimental settings and results will be displayed and analyzed in section \ref{sec:experiments}. Finally, in section \ref{sec:conclusion} there will be a conclusion of our work, followed by a discussion of future work possibilities.

\section{Background}
\label{sec:background}

For the easiness of comprehension, in this section, we briefly cover the theory of deep Gaussian process regression (DGPR). 

DGPR is a regression method based on a multi-layer hierarchical generalization of Gaussian processes and is equivalent to a neural network with multiple hidden layers with infinite width\cite{bui2016deep}. The output of each layer in DGPR is a Gaussian distribution with mean and covariance functions given by Gaussian process modeling. For a test input $\bm x^{*}$, a forward procedure is established by sampling a random variable $ \varepsilon_{l} \sim \mathcal{N} (0,1) $ for each layer and computing the layer output by

\begin{equation}
    \widehat{f}_{l}=\widehat{\boldsymbol{\mu}}_{l}\left(\widehat{f}_{l-1}\right)+\varepsilon_{l} \sqrt{\widehat{\bm{K}}_{l}\left(\widehat{f}_{l-1}, \widehat{f}_{l-1}\right)},
\end{equation}

\noindent
where the hatted variables with subscripts $l$ refer to the estimated functions ($\bm{K}$ for covariance kernel, $\mu$ for mean, $f$ for distribution) of layer $l$.

Similar to the impressive attributes of neural networks in modeling large complicated systems, we expect DGPR a better ability to learn the intrinsic complexity in time series prediction compared to GPR. Previous work\cite{bui2016deep} exploring these two methods provides support to this assumption, and the comparison of them in our experiments will be presented later in section \ref{sec:experiments}.

\section{Methodology}
\label{sec:methodology}

\subsection{Problem Definition}

In a typical battery health prediction problem, we are given a labelled training set of input-output pairs $D = \{(\bm{x_i},y_i)\}_{i=1}^{N_D}$, and the objective is the prediction of $\{y_i\}_{i=N_D}^{N}$ where input $\bm{x_i} \in \mathbb{R}^{d}$ and output $y_i\in \mathbb{R}$. Here $i$ denotes the cycle number; $N_D$ and $N_T=N-N_D$ are the numbers of past (known) and future (to be predicted) data respectively. The output data $y_i$ refers to LIB capacity, and input data $\bm{x_i}$ is a matrix concatenated from profiling temporal information including voltage, temperature and current from every cycle given by BMS. The evaluation metrics are defined as follows.
 
 \begin{equation}
MSE(\hat{y_i},y_i) =\frac{1}{N_T}\Sigma_{i=N_D}^{N}(\hat{y_i} - y_i)^2, 
\end{equation}
 \begin{equation}
R^2(\hat{y_i},y_i) =1-\frac{\Sigma_{i=N_D}^{N}(\hat{y_i} - y_i)^2}{\Sigma_{i=N_D}^{N}(y_i-\frac{1}{N_T}\Sigma_{i=N_D}^{N}y_i)^2}. 
\end{equation}
 
 The notation $\hat{y_i}$ is the predicted capacity value given by the proposed model and $y_i$ is the ground truth capacity value given by measurement. The performance of our proposed method will be evaluated both qualitatively by time evolution graphs and quantitatively by the MSE and $R^2$-score values. For simplicity, the notation $\{\cdot\}$ will be used as an abbreviation for $\{\cdot\}_{i=1}^{N_D}$ in sections below.

\begin{table*}[htb]
\scriptsize
\centering
\caption{MSE and $R^2$-score for different cells and algorithms in battery capacity prediction}
\setlength{\tabcolsep}{3.5mm}{
\begin{tabular}{ccccccccccc}
 \toprule
\multirow{2}{*}{\textbf{Method}} & \multicolumn{2}{c}{\textbf{Cell B0018}} &
\multicolumn{2}{c}{\textbf{Cell B0005}} &
\multicolumn{2}{c}{\textbf{Cell B0006}} &
\multicolumn{2}{c}{\textbf{Cell B0007}} & \multicolumn{2}{c}{\textbf{Avg.}}\\
\cmidrule(r){2-3} \cmidrule(r){4-5} \cmidrule(r){6-7} \cmidrule(r){8-9} \cmidrule(r){10-11}
& \textbf{\textit{MSE}} & \textbf{\textit{$R^2$-score}}& \textbf{\textit{MSE}} & \textbf{\textit{$R^2$-score}} &
\textbf{\textit{MSE}} & \textbf{\textit{$R^2$-score}} & 
\textbf{\textit{MSE}} & \textbf{\textit{$R^2$-score}} & 
\textbf{\textit{MSE}} & \textbf{\textit{$R^2$-score}} \\
 \midrule

DecisionTree & 0.00375 & 0.97493 & 0.00401 & 0.97215 & 0.01147 & 0.95455 & 0.00315 & 0.96933 & 0.00559 & 0.96774\\
XGBoost & 0.00290 & 0.98058 & 0.00413 & 0.97132 & 0.01171 & 0.95362 & 0.00328 & 0.96806 & 0.00550 & 0.96839\\
MLP & 0.00599 & 0.95991 & 0.00170 & 0.98819 & 0.00062 & 0.99755 & 0.00133 & 0.98708 & 0.00241 & 0.98318\\
CNN & 0.00332 & 0.97777 & 0.00181 & 0.98658 & 0.00074 & 0.99634 & 0.00102 & 0.98948 & 0.00172 & 0.98754\\
 \midrule
GPR & 0.00609 & 0.93902 & 0.00118 & 0.98734 & 0.00931 & 0.95390 & 0.00162 & 0.97863 & 0.00455 & 0.96472\\
DGPR & 0.00308 & 0.97935 & 0.00115 & 0.99197 & 0.00100 & 0.99603 & 0.00109 & 0.98936 & 0.00158 & 0.98917\\
LSTM & 0.00779 & 0.94787 & 0.00128 & 0.99111 & 0.00952 & 0.96227 & 0.00198 & 0.98074 & 0.00514 & 0.97049\\

SDG-L w/o EMF & 0.00357 & 0.91888 & 0.02068 & 0.85652 & 0.02343 & 0.90722 & 0.01395 & 0.86454 & 0.01540 & 0.88679 \\

 \midrule

\textbf{SDG-L} & \textbf{0.00250} & \textbf{0.98326} & \textbf{0.00096} & \textbf{0.99329} & \textbf{0.00042} &
\textbf{0.99833} & \textbf{0.00093} & \textbf{0.99096} & \textbf{0.00120} & \textbf{0.99146}\\

\bottomrule

\end{tabular}
}
\label{tab1}
\end{table*}

\subsection{Method Framework}

The pipeline of our method is shown in Fig. \ref{framework}. For capacity data (Y in the figure), 
we fit the capacity series $\{y_i\}$ to the EMF function,
deriving the estimated parameter $\bm{\theta}$ and its residuals $\{r_i\}$. Then for the input concatenated matrix series $\{\bm{x_i}\}$ (X in the figure), 
an LSTM feature extractor is employed to extract features $\{\bm{f_i}\}$. The LSTM and the DGPR module are trained together to model the residuals $\{r_i\}$ with input features $\{\bm{f_i}\}$. 
Lastly, the predicted capacity of a future cycle $i^*\in[N_D,N]$ will be the sum of $m(i^*;\bm{\hat \theta})$ given by the EMF and the DGPR output $\hat f(\bm x_{i^*})$.

\subsection{Feature Extractor}

During the input concatenation within every cycle, for each of the profiling variables (i.e. voltage, current, and temperature for the BMS in NASA dataset), we take one value every five points in the discharge interval until we have twenty points, aligning the profiling series length to 20 and thus the derived matrix size to ${20\times3}$ for each cycle(Fig. \ref{framework}a). Nevertheless, the computational cost of DGPR grows exponentially with the input size, and the concatenated matrix $\bm{x_i}\in \mathbb{R}^{20\times3}$ is intractable for the DGPR's kernel function to process\cite{damianou2013deep}. Therefore an extractor is crucial to reduce the dimension of $\bm{x_i}$ before feeding it into the DGPR module. Considering the temporal structure of profiling data, LSTM is adopted as the feature extractor in our framework. Specifically, the LSTM cell is stacked twice and the number of nodes for the hidden layer is set to 64. The number of nodes for the input layer is set to 3 to match the input size, and for the output layer, it is set to 2 to avoid the exponential increase in computational space and time. The derived feature $\bm{f_i}\in \mathbb{R}^{2}$ is 3D-plotted in Fig. \ref{feature}.

\subsection{Semiparametric Modeling}

The performance of Gaussian process based regression models is largely constrained by the limited expressivity of its mean functions. Therefore, we follow the semiparametric Gaussian processes proposed in \cite{murphy2012machine} by introducing EMFs to DGPR. An EMF typically contains the expression of prior information of battery capacity degradation nature about the expected model. In LIB capacity prediction, several exponential battery degradation models have been derived and verified in previous studies\cite{xu2016modeling}, of which we select the model with the functional form given by
\begin{equation}
    m(i;\bm{\theta}) = \theta_1+\theta_2 e^{\theta_3 i},
\end{equation}

\noindent
where array $\bm{\theta} = (\theta_1, \theta_2, \theta_3)$ is the parameter. The capacity series is fitted to the EMF function using least squares fitting. A zero-mean residual modeling is then implemented on the basis of EMF fitting to explore the existence of hiding contributions to capacity decline apart from the explicit parametric model. The outputs of the EMF and residual models are added up to get the prediction values. Specifically, the DGPR module is two-layered with constant mean and RBF kernels.

\begin{figure*}[ht]
	\begin{center}
		\centerline{\includegraphics[width=1.9\columnwidth]{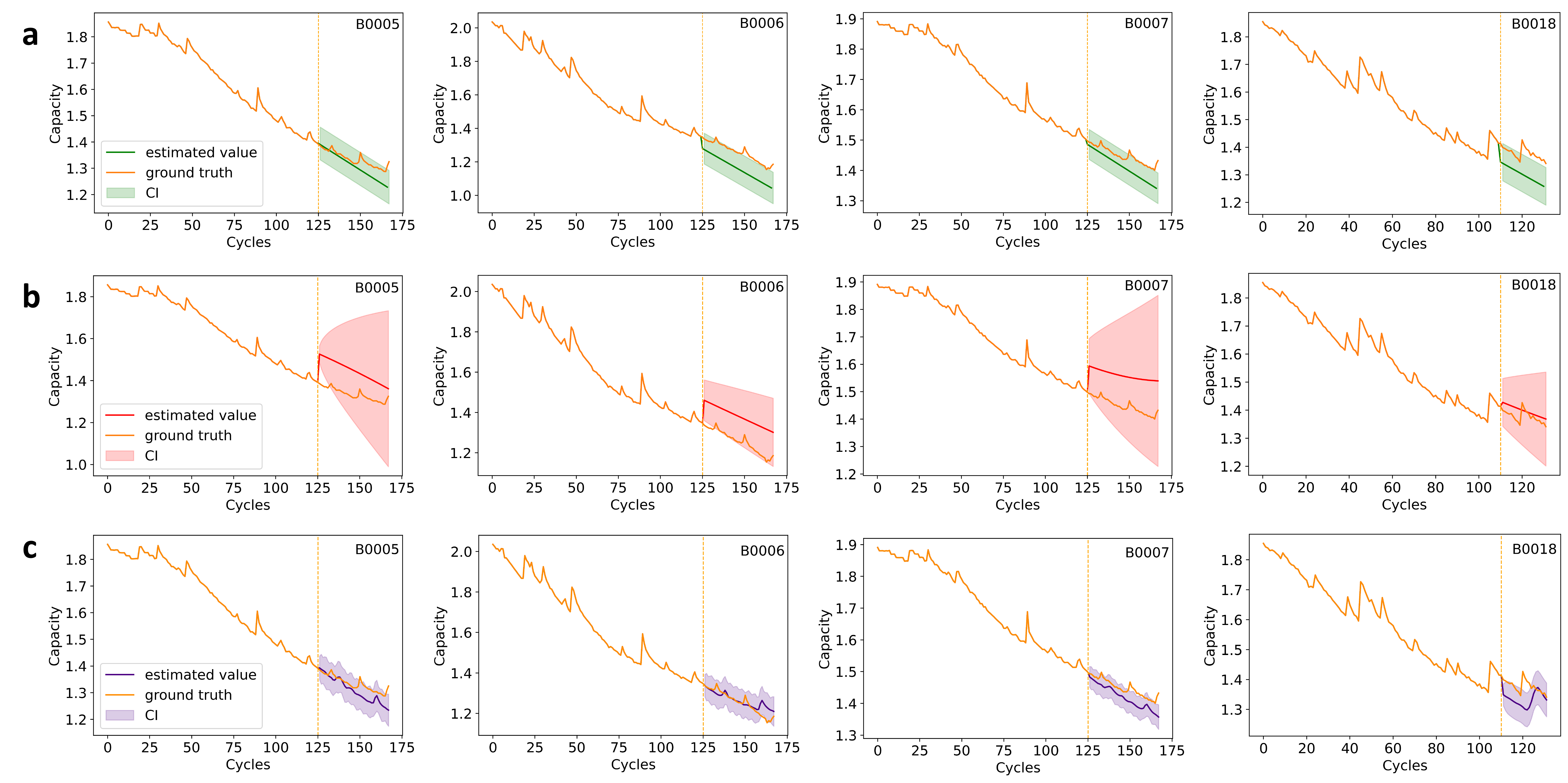}}
		\caption{\textbf{Visualizations of Estimated Value, Ground Truth, and CI.} Orange curves are plotted for the ground truth. Green, red, and purple curves for model prediction of different methods. A 2-sigma CI is marked by translucent regions. The rows from top to bottom (a-c) are the prediction results of GPR, SDG-L w/o curve fitting, and SDG-L; the columns from left to right show LIB cells with ID B0005, B0006, B0007, and B0018. The training number for cells B0005, B0006, and B0007 is set as 125 and B0018 as 110, marked by the vertical dashed line. }
		\label{set1}
	\end{center}
 \vspace{-0.5cm}

\end{figure*}

\section{Experiment Results}
\label{sec:experiments}

\subsection{Data Description}

The battery data we use is obtained from the NASA Ames Prognostics Center of Excellence Battery Dataset \cite{NASA}. By running through three different operational profiles (charge, discharge, and impedance) at room temperature for four lithium-ion batteries (B0005, B0006, B0007, B0018), a set under these three conditions is recorded. We take full account of the time series attributes in every cycle, and get the feature vector for every cycle for these four batteries respectively. As for output, the capacity of a specific cycle can be obtained from the discharge mode of source data which are real data from the above NASA Battery Dataset. 
There are 168 cycles for batteries B0005, B0006, and B0007 and 132 cycles for battery B0018. As the charge and discharge cycle increases, the battery capacity gradually decreases.

\subsection{Prediction Results}

Comprehensive experiments are conducted on SDG-L and other common methods. Traditional methods used for prediction such as Decision Tree, XGBoost, MLP and CNN are all evaluated on the NASA dataset. The effectiveness of SDG-L is verified by quantitative comparison among these methods. The number of training epochs is set to be 200 as enough for the loss value to converge. The learning rate for DGPR training is set to 0.1. The training size $N_D$ for cell B0005, B0006, and B0007 is set to 125 and B0018 to 110. 

\begin{figure}[ht]
	\begin{center}
		\centerline{\includegraphics[width=0.8\columnwidth]{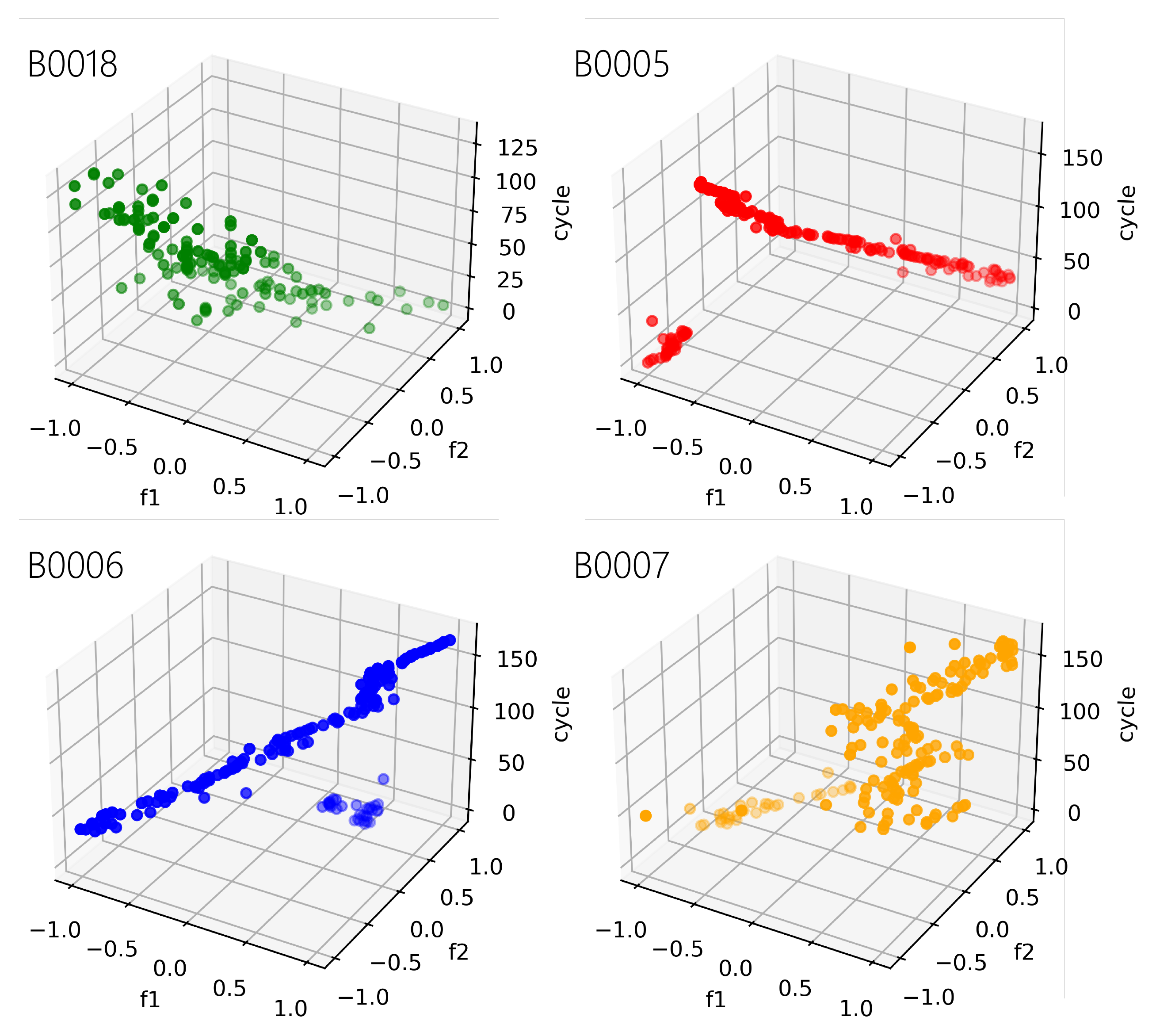}}
		\caption{\textbf{Feature Series Extracted from LSTM Network of Different LIB Cells.} (Upper Left: B0018, Upper Right: B0005, Lower Left: B0006, Lower Right: B0007) We have $x$,$y$ axis for the two-dimensional features and $z$ axis for cycle numbers.}
		\label{feature}
	\end{center}
\vspace{-0.5cm}
\end{figure}

A quantitative evaluation for these methods are presented in the first half of Table \ref{tab1} where our proposed SDG-L (the last row) outperforms the others with the lowest MSE and highest $R^2$-score. The variation of model performance among columns (different cells) results from the different predictability of cells caused by their differed physical properties. Fig. \ref{feature} visualizes the domain difference across them, highlighting the effectiveness of LSTM in capturing such diversity as well as indicating that domain adaptation and generalization of our proposed prediction should be considered in our future work. 

Overall, SDG-L reduces the average error by 78.5\% at most compared with Decision Tree and 30.2\% at least compared with CNN. It is worth noting that the order of results obtained by different methods varies on different battery cells. For example, Decision Tree gives better results than XGBoost on cell B001818, but worse results on cell B0005. However, our method gives the best results on all battery cells which reflects the robustness and efficiency of SDG-L.

\subsection{Ablation Study}

To evaluate the impact and performance of each component in our framework, we evaluate their effectiveness in this section. The results are reported in the bottom half of the Table \ref{tab1} which represents plain GPR, DGPR, LSTM and SDG-L without curve fitting from top to bottom. After replacing GPR with DGPR, the prediction results of all the battery cells are improved with lower MSE and higher $R^2$-score which proves the effectiveness of the DGPR. The results of directly using LSTM to predict battery capacity are shown in the third row which are not satisfactory. However, our proposed SDG-L obtains better results by combining LSTM and DGPR which are shown in the last row. EMF also proves to be indispensable by comparing the last two rows.

The utilization of auxiliary features together with the expressivity and flexibility of LSTM are the main reasons that SDG-L outperforms DGPR. The nonlinear structure and the time correlation attribute of LSTM allow more diverse features to be acquired, and therefore DGPR can utilize more comprehensive information to predict overall iterative processes rather than deterministic features in GPR or time features in DGPR. 

The illustrative prediction performance for each cell is compared among a GPR with a white kernel, SDG-L w/o curve fitting but linear mean, and our proposed SDG-L method framework, which are shown in the rows from top to down in Fig. \ref{set1} (a-c). Our method obtains the best average prediction shown in the last row of Table \ref{tab1} with fluctuations roughly coinciding with the ground truth and a narrow two-sigma confidential interval covering most of the real values. As can be seen in Fig. \ref{set1}c, SDG-L could capture some non-linearity while the other two models tend to produce a linear prediction, which owes to the better expressivity of DGPR on complex modes compared to GPR and the comprehensive information extracted from LSTM. The effectiveness of EMF in battery capacity prediction can be well reflected by comparing the results of SDG-L (Fig. \ref{set1}c) with those of SDG-L w/o EMF (Fig. \ref{set1}b).

\section{Conclusion}
\label{sec:conclusion}

Driven by the growing demand for accurate lithium-ion battery capacity prediction and relative lacking in the utilization of state information during charging/discharging cycles, we propose a semiparametric deep Gaussian process based framework with an LSTM feature extractor, abbreviated as SDG-L. The highly representative DGPR module leverages the temporal features to predict the battery capacity in future cycles. Particularly, the methodology uses LSTM to extract features from BMS profiling information; an EMF is also introduced in combination with DGPR to exploit prior information of battery degradation.
Our proposed SDG-L maximizes the use of all available information and its effectiveness is validated by experiments conducted on NASA battery dataset using ablation studies. We also show in qualitative and quantitative comparison that our algorithm significantly outperforms commonly used methods like decision tree, XGBoost, MLP and CNN.

In future works, we will consider changing the framework to an end-to-end one cascading curve fitting with module training. Also, the robustness of our framework can be improved by handling domain shift in battery data and conducting multi-task learning with other related LIB parameters such as state of function (SoF) and state of charge (SoC).

\section*{Acknowledgements}

This study is supported by the Tsinghua SIGS Scientific Research Start-up Fund (Grant No. QD2021012C), Shenzhen Science and Technology Innovation Commission (Grant JCYJ20220530143002005), High-end Foreign Expert Talent Introduction Plan (G2021032021L), Shenzhen Science and Technology Program (Grant No. KQTD20170810150821146) and Tsinghua Shenzhen International  Graduate  School  Interdisciplinary  Innovative  Fund (JC2021006).

\bibliographystyle{IEEEtran}
\bibliography{mybibfile}

\end{document}